\pdfoutput=1
\documentclass[10pt,twocolumn,letterpaper]{article}

\usepackage[pagenumbers]{cvpr} %
\makeatletter
\patchcmd{\@maketitle}{\vspace*{24pt}}{\vspace*{10.06pt}}{}{\PackageWarning{main}{Could not patch the title spacing}}
\makeatother

\usepackage{xspace}
\usepackage{algorithm}
\usepackage{algpseudocode}
\usepackage{listings}
\usepackage{multirow}
\usepackage{colortbl}
\usepackage{tikz}
\usetikzlibrary{arrows.meta,positioning}

\definecolor{seedbg}{RGB}{232,242,252}
\definecolor{twiffbg}{RGB}{253,239,220}

\newcommand{\method}{Future-OPSD\xspace}

\newcommand{\cmark}{\ensuremath{\checkmark}}

\newcommand{\KL}{D_{\mathrm{KL}}}
\newcommand{\E}{\mathbb{E}}

\newcommand{\Lc}{\mathcal L}
\newcommand{\Rc}{\mathcal R}
\newcommand{\zt}{\tilde z}
\newcommand{\ind}{\mathrel{\perp\!\!\!\perp}}
\DeclareMathOperator{\clip}{clip}

\usepackage{amsthm}
\newtheorem{theorem}{Theorem}

\AtEndPreamble{%
  \crefname{assumption}{Assumption}{Assumptions}%
  \Crefname{assumption}{Assumption}{Assumptions}%
}

\makeatletter
\newbox\opd@pendinglabels
\@ifundefined{@ShipoutPicture}{}{%
  \let\opd@cvprShipoutPicture\@ShipoutPicture
  \def\@ShipoutPicture{%
    \global\setbox\opd@pendinglabels\box\@labels
    \let\opd@ifinlabel\if@inlabel
    \global\@inlabelfalse
    \opd@cvprShipoutPicture
    \global\setbox\@labels\box\opd@pendinglabels
    \global\let\if@inlabel\opd@ifinlabel}%
}
\makeatother

\providecommand{\msd}[2]{#1}
\providecommand{\msdimage}[2]{#1}
\providecommand{\basedelta}[1]{\textcolor{blue}{#1}}
\providecommand{\swatch}[1]{{\setlength{\fboxsep}{0pt}\setlength{\fboxrule}{0.4pt}\fcolorbox{black!50}{#1}{\rule{0pt}{0.7em}\rule{0.7em}{0pt}}}}

\definecolor{cvprblue}{rgb}{0.21,0.49,0.74}
\usepackage{xr-hyper}
\usepackage[pagebackref,breaklinks,colorlinks,allcolors=cvprblue]{hyperref}
\AtBeginDocument{\makeatletter\expandafter\gdef\csname r@app:proof\endcsname{{A}{1}{\texorpdfstring {\hskip -1em.~}{}Modeling and Derivations}{appendix.A}{supp.pdf}}
\expandafter\gdef\csname r@app:proof@cref\endcsname{{[appendix][1][2147483647]A}{[1][1][]1}}
\expandafter\gdef\csname r@app:setup\endcsname{{A.1}{1}{\texorpdfstring {\hskip -1em.~}{}Modeling and Notation}{subsection.A.1}{supp.pdf}}
\expandafter\gdef\csname r@app:setup@cref\endcsname{{[subappendix][1][2147483647,1]A.1}{[1][1][]1}}
\expandafter\gdef\csname r@app:notation\endcsname{{A.1}{1}{\texorpdfstring {\hskip -1em.~}{}Modeling and Notation}{subsection.A.1}{supp.pdf}}
\expandafter\gdef\csname r@app:notation@cref\endcsname{{[subappendix][1][2147483647,1]A.1}{[1][1][]1}}
\expandafter\gdef\csname r@eq:joint\endcsname{{A.1}{1}{Sampling and information available to the student}{equation.A.1}{supp.pdf}}
\expandafter\gdef\csname r@eq:joint@cref\endcsname{{[equation][1][2147483647,1]A.1}{[1][1][]1}}
\expandafter\gdef\csname r@eq:app-f\endcsname{{A.2}{1}{Teacher scoring}{equation.A.2}{supp.pdf}}
\expandafter\gdef\csname r@eq:app-f@cref\endcsname{{[equation][2][2147483647,1]A.2}{[1][1][]1}}
\expandafter\gdef\csname r@app:identity\endcsname{{A.2}{1}{\texorpdfstring {\hskip -1em.~}{}Proof of Theorem 1}{subsection.A.2}{supp.pdf}}
\expandafter\gdef\csname r@app:identity@cref\endcsname{{[subappendix][2][2147483647,1]A.2}{[1][1][]1}}
\expandafter\gdef\csname r@eq:q\endcsname{{A.3}{1}{\texorpdfstring {\hskip -1em.~}{}Proof of Theorem 1}{equation.A.3}{supp.pdf}}
\expandafter\gdef\csname r@eq:q@cref\endcsname{{[equation][3][2147483647,1]A.3}{[1][1][]1}}
\expandafter\gdef\csname r@eq:finite-nce\endcsname{{A.4}{2}{Connection to InfoNCE}{equation.A.4}{supp.pdf}}
\expandafter\gdef\csname r@eq:finite-nce@cref\endcsname{{[equation][4][2147483647,1]A.4}{[1][1][]2}}
\expandafter\gdef\csname r@app:certificate\endcsname{{A.3}{2}{\texorpdfstring {\hskip -1em.~}{}Proof of Theorem 2}{subsection.A.3}{supp.pdf}}
\expandafter\gdef\csname r@app:certificate@cref\endcsname{{[subappendix][3][2147483647,1]A.3}{[1][2][]2}}
\expandafter\gdef\csname r@eq:app-contrast\endcsname{{A.5}{2}{From the foresight score to an additive contrast}{equation.A.5}{supp.pdf}}
\expandafter\gdef\csname r@eq:app-contrast@cref\endcsname{{[equation][5][2147483647,1]A.5}{[1][2][]2}}
\expandafter\gdef\csname r@eq:psi\endcsname{{A.6}{2}{From the foresight score to an additive contrast}{equation.A.6}{supp.pdf}}
\expandafter\gdef\csname r@eq:psi@cref\endcsname{{[equation][6][2147483647,1]A.6}{[1][2][]2}}
\expandafter\gdef\csname r@eq:app-gap\endcsname{{A.7}{2}{From the foresight score to an additive contrast}{equation.A.7}{supp.pdf}}
\expandafter\gdef\csname r@eq:app-gap@cref\endcsname{{[equation][7][2147483647,1]A.7}{[1][2][]2}}
\expandafter\gdef\csname r@eq:C\endcsname{{A.8}{2}{From the foresight score to an additive contrast}{equation.A.8}{supp.pdf}}
\expandafter\gdef\csname r@eq:C@cref\endcsname{{[equation][8][2147483647,1]A.8}{[1][2][]2}}
\expandafter\gdef\csname r@eq:app-bound\endcsname{{A.9}{2}{From the foresight score to an additive contrast}{equation.A.9}{supp.pdf}}
\expandafter\gdef\csname r@eq:app-bound@cref\endcsname{{[equation][9][2147483647,1]A.9}{[1][2][]2}}
\expandafter\gdef\csname r@app:advantage\endcsname{{A.4}{2}{\texorpdfstring {\hskip -1em.~}{}From the Sequence Objective to Token Updates}{subsection.A.4}{supp.pdf}}
\expandafter\gdef\csname r@app:advantage@cref\endcsname{{[subappendix][4][2147483647,1]A.4}{[1][2][]2}}
\expandafter\gdef\csname r@eq:app-token-decomposition\endcsname{{A.10}{2}{Sequence decomposition}{equation.A.10}{supp.pdf}}
\expandafter\gdef\csname r@eq:app-token-decomposition@cref\endcsname{{[equation][10][2147483647,1]A.10}{[1][2][]2}}
\expandafter\gdef\csname r@eq:app-token-reward\endcsname{{A.11}{2}{Sequence decomposition}{equation.A.11}{supp.pdf}}
\expandafter\gdef\csname r@eq:app-token-reward@cref\endcsname{{[equation][11][2147483647,1]A.11}{[1][2][]2}}
\expandafter\gdef\csname r@eq:app-sequence-gradient\endcsname{{A.12}{2}{Exact gradient and the immediate approximation}{equation.A.12}{supp.pdf}}
\expandafter\gdef\csname r@eq:app-sequence-gradient@cref\endcsname{{[equation][12][2147483647,1]A.12}{[1][2][]2}}
\expandafter\gdef\csname r@eq:app-immediate-advantage\endcsname{{A.13}{3}{Exact gradient and the immediate approximation}{equation.A.13}{supp.pdf}}
\expandafter\gdef\csname r@eq:app-immediate-advantage@cref\endcsname{{[equation][13][2147483647,1]A.13}{[1][3][]3}}
\expandafter\gdef\csname r@app:algorithm\endcsname{{A.5}{3}{\texorpdfstring {\hskip -1em.~}{}Algorithm Flowchart and Pseudocode}{subsection.A.5}{supp.pdf}}
\expandafter\gdef\csname r@app:algorithm@cref\endcsname{{[subappendix][5][2147483647,1]A.5}{[1][3][]3}}
\expandafter\gdef\csname r@fig:training-flow\endcsname{{S1}{3}{\textbf {One training iteration of \method .} Dashed arrows carry futures only to the frozen teacher. Both teacher views score the same student response without generating a new one. Only valid response tokens enter advantage processing and PPO.\relax }{figure.caption.13}{supp.pdf}}
\expandafter\gdef\csname r@fig:training-flow@cref\endcsname{{[figure][1][2147483647]S1}{[1][3][]3}}
\expandafter\gdef\csname r@alg:future-opsd\endcsname{{1}{4}{\method training\relax }{algorithm.1}{supp.pdf}}
\expandafter\gdef\csname r@alg:future-opsd@cref\endcsname{{[algorithm][1][2147483647]1}{[1][3][]4}}
\expandafter\gdef\csname r@app:additional\endcsname{{B}{4}{\texorpdfstring {\hskip -1em.~}{}Additional Experimental Results}{appendix.B}{supp.pdf}}
\expandafter\gdef\csname r@app:additional@cref\endcsname{{[appendix][2][2147483647]B}{[1][4][]4}}
\expandafter\gdef\csname r@app:general-evaluation\endcsname{{B.1}{4}{\texorpdfstring {\hskip -1em.~}{}General Video Understanding Evaluation}{subsection.B.1}{supp.pdf}}
\expandafter\gdef\csname r@app:general-evaluation@cref\endcsname{{[subappendix][1][2147483647,2]B.1}{[1][4][]4}}
\expandafter\gdef\csname r@app:future-sampling\endcsname{{B.2}{4}{\texorpdfstring {\hskip -1em.~}{}Future Frame Sampling}{subsection.B.2}{supp.pdf}}
\expandafter\gdef\csname r@app:future-sampling@cref\endcsname{{[subappendix][2][2147483647,2]B.2}{[1][4][]4}}
\expandafter\gdef\csname r@sec:opd\endcsname{{B.3}{4}{\texorpdfstring {\hskip -1em.~}{}Extension to Teacher--Student OPD}{subsection.B.3}{supp.pdf}}
\expandafter\gdef\csname r@sec:opd@cref\endcsname{{[subappendix][3][2147483647,2]B.3}{[1][4][]4}}
\expandafter\gdef\csname r@app:datascale-ego\endcsname{{B.4}{4}{\texorpdfstring {\hskip -1em.~}{}Scaling on EgoPlan-IT}{subsection.B.4}{supp.pdf}}
\expandafter\gdef\csname r@app:datascale-ego@cref\endcsname{{[subappendix][4][2147483647,2]B.4}{[1][4][]4}}
\expandafter\gdef\csname r@sec:datascale\endcsname{{B.4}{4}{\texorpdfstring {\hskip -1em.~}{}Scaling on EgoPlan-IT}{subsection.B.4}{supp.pdf}}
\expandafter\gdef\csname r@sec:datascale@cref\endcsname{{[subappendix][4][2147483647,2]B.4}{[1][4][]4}}
\expandafter\gdef\csname r@tab:q2b-egoplan\endcsname{{S1}{5}{\textbf {Egocentric planning (Qwen3.5-2B).} Trained on EgoPlan-IT; distillation uses future frames ($z=\text {image}$). Single runs; bold marks column maxima. Blue $\Delta $ is the preceding method minus Base.\relax }{table.caption.18}{supp.pdf}}
\expandafter\gdef\csname r@tab:q2b-egoplan@cref\endcsname{{[table][1][2147483647]S1}{[1][4][]5}}
\expandafter\gdef\csname r@tab:opd-q35\endcsname{{S2}{5}{\textbf {Extension to teacher--student OPD with Qwen3.5-4B on EgoPlan-IT.} The student uses its frozen initialization for OPSD or a frozen Qwen3.5-9B teacher for OPD; formatting follows \cref {tab:opd}. Blue $\Delta $ rows report the preceding image-based method minus Base.\relax }{table.caption.19}{supp.pdf}}
\expandafter\gdef\csname r@tab:opd-q35@cref\endcsname{{[table][2][2147483647]S2}{[1][4][]5}}
\expandafter\gdef\csname r@fig:datascale\endcsname{{S2}{5}{\textbf {Scaling on EgoPlan-IT.} SEED-Bench-R1 average accuracy of Qwen3-VL-8B as the number of training prompts $N$ increases; hatched bars ($^\dagger $) train with answers, and the dotted line marks the base model. GRPO samples eight rollouts per prompt and runs one-eighth of the steps (on the first $N/8$ prompts), generating as many responses as OPSD. Green numbers give the gain of Future-OPSD over Naive OPSD; the values are listed in \cref {tab:datascale}.\relax }{figure.caption.20}{supp.pdf}}
\expandafter\gdef\csname r@fig:datascale@cref\endcsname{{[figure][2][2147483647]S2}{[1][4][]5}}
\expandafter\gdef\csname r@tab:datascale\endcsname{{S3}{5}{\textbf {Scaling on EgoPlan-IT.} SEED-Bench-R1 average accuracy as the number of training prompts increases. $\Delta $ is Future-OPSD minus Naive OPSD; $^\dagger $ denotes training with answers. GRPO samples eight rollouts per prompt and runs one-eighth of the steps (on the first $N/8$ prompts), generating as many responses as OPSD. Bold marks the highest score at each data scale, including ties.\relax }{table.caption.22}{supp.pdf}}
\expandafter\gdef\csname r@tab:datascale@cref\endcsname{{[table][3][2147483647]S3}{[1][5][]5}}
\expandafter\gdef\csname r@app:datascale-twiff\endcsname{{B.5}{5}{\texorpdfstring {\hskip -1em.~}{}Scaling on Future-L1}{subsection.B.5}{supp.pdf}}
\expandafter\gdef\csname r@app:datascale-twiff@cref\endcsname{{[subappendix][5][2147483647,2]B.5}{[1][5][]5}}
\expandafter\gdef\csname r@tab:datascale-q3vl2b\endcsname{{S4}{6}{\textbf {Scaling on EgoPlan-IT (Qwen3-VL-2B-Instruct).} SEED-Bench-R1 average accuracy as the number of training prompts increases, under the protocol of \cref {tab:datascale}. $\Delta $ is Future-OPSD minus Naive OPSD; $^\dagger $ denotes training with answers. SFT uses the 2B model's own samples with an enforced output format. Naive OPSD and Future-OPSD use the student itself as the frozen teacher, and Future-OPD$^\ddagger $ a frozen Qwen3-VL-4B-Instruct teacher. All distillation variants use future frames ($z=\text {image}$); the contrastive variants use $\lambda =0.75$. Unlike \cref {tab:datascale}, GRPO follows \cref {tab:q3vl-egoplan}: 32 queries $\times $ 8 rollouts per step with one policy update per step, for as many steps as OPSD (on the first $N/8$ prompts). Bold marks the highest score at each data scale, including ties.\relax }{table.caption.23}{supp.pdf}}
\expandafter\gdef\csname r@tab:datascale-q3vl2b@cref\endcsname{{[table][4][2147483647]S4}{[1][5][]6}}
\expandafter\gdef\csname r@tab:datascale-twiff\endcsname{{S5}{6}{\textbf {Scaling on Future-L1.} TwiFF-Bench average score as the number of training prompts increases; $\Delta $ is Future-OPSD minus Naive OPSD. Bold marks the highest score at each data scale, including ties.\relax }{table.caption.24}{supp.pdf}}
\expandafter\gdef\csname r@tab:datascale-twiff@cref\endcsname{{[table][5][2147483647]S5}{[1][5][]6}}
\expandafter\gdef\csname r@app:ablation-values\endcsname{{B.6}{6}{\texorpdfstring {\hskip -1em.~}{}Ablation Values}{subsection.B.6}{supp.pdf}}
\expandafter\gdef\csname r@app:ablation-values@cref\endcsname{{[subappendix][6][2147483647,2]B.6}{[1][6][]6}}
\expandafter\gdef\csname r@app:training-cost\endcsname{{B.7}{6}{\texorpdfstring {\hskip -1em.~}{}Detailed Training Costs}{subsection.B.7}{supp.pdf}}
\expandafter\gdef\csname r@app:training-cost@cref\endcsname{{[subappendix][7][2147483647,2]B.7}{[1][6][]6}}
\expandafter\gdef\csname r@tab:privilege-egoplan\endcsname{{S6}{6}{\textbf {Text versus image privilege on EgoPlan-IT.} In-domain SEED-Bench-R1 accuracy (\%). Future-OPSD rows report three-seed means; Naive OPSD uses single runs. Bold marks column maxima within each model.\relax }{table.caption.25}{supp.pdf}}
\expandafter\gdef\csname r@tab:privilege-egoplan@cref\endcsname{{[table][6][2147483647]S6}{[1][6][]6}}
\expandafter\gdef\csname r@tab:lambda\endcsname{{S7}{6}{\textbf {Mixing weight $\lambda $} (\cref {eq:cpsd}), evaluated on the training domain: SEED-Bench-R1 average for EgoPlan-IT and TwiFF-Bench for Future-L1. $\lambda =0$ is Naive OPSD with future frames, trained with the same recipe. $\lambda =0.5$ and the default $\lambda =0.75$ are means over three seeds; all other values are single runs. Bold marks the best result for each domain--student pair. $^\dagger $Training collapsed.\relax }{table.caption.26}{supp.pdf}}
\expandafter\gdef\csname r@tab:lambda@cref\endcsname{{[table][7][2147483647]S7}{[1][6][]6}}
\expandafter\gdef\csname r@tab:teacher-vit\endcsname{{S8}{6}{\textbf {Teacher weights and vision encoder} for Future-OPSD ($z=\text {image}$, $\lambda =0.75$), with Qwen3.5-2B trained on EgoPlan-IT, evaluated on SEED-Bench-R1 (in-domain). All rows below Base are Future-OPSD; the first is the default recipe (teacher frozen at the initial model, vision encoder trained); each other row changes one setting. The EMA teacher is updated after every step with decay 0.98. Bold marks the highest score in each column (single seed).\relax }{table.caption.28}{supp.pdf}}
\expandafter\gdef\csname r@tab:teacher-vit@cref\endcsname{{[table][8][2147483647]S8}{[1][6][]6}}
\expandafter\gdef\csname r@tab:teacher-vit-fl1\endcsname{{S9}{7}{\textbf {Teacher weights and vision encoder} for Future-OPSD ($z=\text {image}$, $\lambda =0.75$), with Qwen3-VL-2B-Instruct trained on Future-L1, evaluated on TwiFF-Bench (in-domain). \textcolor {red}{Invalid results, discarded.} All rows below Base are Future-OPSD; the first is the default recipe (teacher frozen at the initial model, vision encoder trained); each other row changes one setting. The EMA teacher is updated after every step with decay 0.98. Bold marks the highest score in each column (single seed).\relax }{table.caption.29}{supp.pdf}}
\expandafter\gdef\csname r@tab:teacher-vit-fl1@cref\endcsname{{[table][9][2147483647]S9}{[1][6][]7}}
\expandafter\gdef\csname r@tab:cost-egoplan-detail\endcsname{{S10}{7}{\textbf {Archived training cost} for Qwen3-VL-8B on EgoPlan-IT (100 steps; 8 B200 GPUs). \emph {Time} is wall-clock/GPU-hours; $Q\times R$ is queries per step $\times $ rollouts per query; T-pass is frozen-teacher forward passes per rollout. SFT excludes reasoning-trace generation. Bold marks the highest SEED-Bench-R1 accuracy.\relax }{table.caption.33}{supp.pdf}}
\expandafter\gdef\csname r@tab:cost-egoplan-detail@cref\endcsname{{[table][10][2147483647]S10}{[1][7][]7}}
\expandafter\gdef\csname r@app:controls\endcsname{{B.8}{7}{\texorpdfstring {\hskip -1em.~}{}Additional Controls and Diagnostics}{subsection.B.8}{supp.pdf}}
\expandafter\gdef\csname r@app:controls@cref\endcsname{{[subappendix][8][2147483647,2]B.8}{[1][7][]7}}
\expandafter\gdef\csname r@app:futurebench\endcsname{{B.8}{7}{Option bias in FutureBench}{section*.35}{supp.pdf}}
\expandafter\gdef\csname r@app:futurebench@cref\endcsname{{[subappendix][8][2147483647,2]B.8}{[1][7][]7}}
\expandafter\gdef\csname r@tab:cost-futurel1\endcsname{{S11}{8}{\textbf {Archived training cost on Future-L1-50K} (Qwen3-VL-8B; eight GPUs). Time is measured wall-clock hours/GPU-hours; FLOPs are analytical estimates. GRPO includes its 27B judge. SFT trains for 391 steps (one epoch); the other methods train for 100 steps.\relax }{table.caption.34}{supp.pdf}}
\expandafter\gdef\csname r@tab:cost-futurel1@cref\endcsname{{[table][11][2147483647]S11}{[1][7][]8}}
\expandafter\gdef\csname r@app:cases\endcsname{{C}{9}{\texorpdfstring {\hskip -1em.~}{}Case Analysis}{appendix.C}{supp.pdf}}
\expandafter\gdef\csname r@app:cases@cref\endcsname{{[appendix][3][2147483647]C}{[1][9][]9}}
\expandafter\gdef\csname r@fig:cases\endcsname{{S3}{10}{\textbf {Case studies on TwiFF-Bench.} Qwen3-VL-8B before training (Base) and after training on Future-L1 with Naive OPSD ($z=\text {image}$) and with Future-OPSD ($z$=image). The left column gives the question, the observed frame that every model receives, and the future clip, which no model sees. The other columns quote each model's reasoning (verbatim; \ldots \ marks omissions) and give its full answer; the boxes give the judge's two scores (0--5), one for the reasoning (CoT) and one for the answer (Ans.). \textcolor {casebad}{Red}: a generic script, a refusal, or a claim about frames the model was not given; \textcolor {casegood}{green}: a cue in the observed frame that the prediction relies on. The base model falls back on a generic script (rows 1--2) or declines to predict (row 3). Naive OPSD keeps these answers, even where it claims to see the next frames (rows 2--3). Future-OPSD predicts the step that the observed frame points to: the bowl of vegetables next to the hand, the raw meat that still has to be mixed, and the yellow labels to the right of the soda water. The cases are drawn from the 92 of 1,078 questions on which Future-OPSD's answer scores at least 4 and the other two answers at most 2.\relax }{figure.caption.37}{supp.pdf}}
\expandafter\gdef\csname r@fig:cases@cref\endcsname{{[figure][3][2147483647]S3}{[1][9][]10}}
\expandafter\gdef\csname r@fig:cases-seed\endcsname{{S4}{11}{\textbf {Case studies on SEED-Bench-R1} (Qwen3-VL-8B trained on EgoPlan-IT). The left column gives the question and its four options ($\checkmark $: correct), and three frames of the progress video and the current observation that every model receives. It also lists the steps that the dataset annotates in the progress video; the models do not see this annotation. The other columns quote each model's full reasoning (the text inside its \texttt {<reason>} tags) and give its choice; the models output only the letter, and we add the option text. \textcolor {casebad}{Red}: a step that is already done, and the reasoning that leads to it; \textcolor {casegood}{green}: a cue in the current observation that the prediction relies on. The base model and Naive OPSD repeat a completed step: the cupboard is already open, the dryer has been opened, loaded, and closed, and the brush has been rinsed and put down. Future-OPSD instead reads the state of the current observation: the open cupboard, the hands at the coin slot, and the lizard lifted out of the bowl next to the towel.\relax }{figure.caption.38}{supp.pdf}}
\expandafter\gdef\csname r@fig:cases-seed@cref\endcsname{{[figure][4][2147483647]S4}{[1][9][]11}}
\makeatother}
\hypersetup{
  pdfsubject={arXiv preprint},
  pdfauthor={Jingchen Sun, Jiaxing Zhang, Qingqiao Hu, Dingqiang Ye, Yuqi Jia, Yijun Liang, Di Fu, Yunjie Tian, Dongdi Zhao, Changyou Chen}
}

\def\confName{CVPR}
\def\confYear{2027}

\title{Learning Video Foresight Reasoning from Unlabeled Videos\texorpdfstring{\\}{ }via On-Policy Self-Distillation}
\hypersetup{pdftitle={\thetitle}}
\author{
  Jingchen Sun\textsuperscript{1*} \quad Jiaxing Zhang\textsuperscript{*} \quad Qingqiao Hu\textsuperscript{2} \quad Dingqiang Ye\textsuperscript{3} \quad Yuqi Jia\textsuperscript{4}\\
  Yijun Liang\textsuperscript{5} \quad Di Fu \quad Yunjie Tian \quad Dongdi Zhao\textsuperscript{\dag} \quad Changyou Chen\textsuperscript{1}\\[0.3em]
  {\normalsize \textsuperscript{1}University at Buffalo \quad \textsuperscript{2}Stony Brook University \quad \textsuperscript{3}Johns Hopkins University}\\
  {\normalsize \textsuperscript{4}Duke University \quad \textsuperscript{5}University of Maryland, College Park}
}

\begin{document}
\maketitle
{\renewcommand{\thefootnote}{}\footnotetext{\textsuperscript{*}Equal contribution. \quad \textsuperscript{\dag}Corresponding author.}}
\begin{abstract}
Video foresight reasoning requires vision-language models to anticipate future events or next actions from currently observed videos.
Existing methods fit future captions, rationales, or full visual states, but face two limitations.
First, they rely on annotations or external large models for supervision, limiting scalability.
We address this with an on-policy self-distillation framework that turns past--future video pairs into a natural self-supervised signal without extra annotations or LLM supervision.
Second, they treat the whole future clip as equally valuable, although most of it is irrelevant context information.
We address this by defining a \emph{foresight score}, from which we obtain a contrastive privilege advantage that drives the model to learn the most valuable future information.
Since the score is intractable, we optimize a tractable lower-bound surrogate that interpolates this advantage, computed between the true and a random reference future, with a teacher--student distillation advantage; together, they amplify decision-relevant content information and suppress irrelevant context information.
We call the resulting framework \method.
On open-ended prediction and closed-ended planning, \method improves base models by up to 14.2 and 15.9 points without compromising general video understanding.
Code will be available at \href{https://jingchensun.github.io/future-opsd/}{https://jingchensun.github.io/future-opsd/}.
\end{abstract}

\section{Introduction}
\label{sec:intro}

\suppressfloats[t]
\begin{figure}[t]
  \centering
  \includegraphics[width=\linewidth]{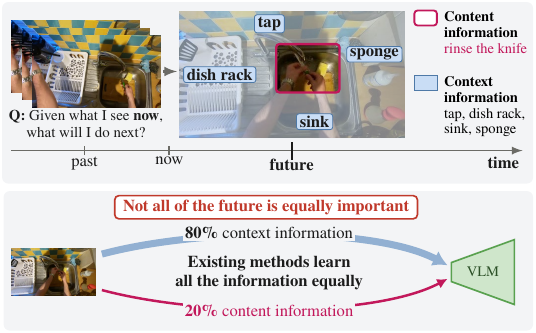}
  \caption{\textbf{Not all of the future is equally important.}
To answer \emph{what will I do next?} from what is seen now, only a small part of the future clip carries content information (\emph{rinse the knife}); most of it is context information (tap, dish rack, sink, sponge) that neither answers the question nor reflects the expected action and trajectory of the ongoing event.}
  \label{fig:teaser}
  \vspace{-12pt}
\end{figure}

\begingroup
\color{black}
Video foresight reasoning aims to infer plausible future events or
states from an observed video by reasoning over its visual dynamics,
causal relations, and agent intentions.
This ability has been studied through diverse task formulations,
including activity forecasting and action
anticipation~\cite{kitani2012activity,damen2018scaling,grauman2022ego4d,furnari2019rulstm,girdhar2021anticipative,zhao2024antgpt},
trajectory forecasting~\cite{alahi2016social,gupta2018social,salzmann2020trajectron},
future frame and representation
prediction~\cite{vondrick2016anticipating,mathieu2016deep,finn2016unsupervised},
and video-language future event prediction~\cite{yi2020clevrer,lei2020futureevent}.
With the advent of vision-language models (VLMs), recent work recasts
foresight as high-level reasoning, with dedicated
benchmarks~\cite{gong2026thinking,chen2026futureomni,shi2026river,lan2026fevos}
and reasoning-oriented training.
Foresight underpins world models for planning and
control~\cite{ha2018recurrent,hafner2020dream,du2023unipi,assran2025vjepa2},
and is essential for autonomous driving~\cite{hu2023gaia1}, embodied
robotics~\cite{finn2016unsupervised,du2023unipi,assran2025vjepa2}, and
proactive egocentric assistants~\cite{damen2018scaling,grauman2022ego4d}.

Recent VLM-based methods have begun to explicitly train foresight
reasoning.
Merlin~\cite{yu2024merlin} injects foresight through trajectory-based
pre-training and instruction tuning.
Next-Event Prediction~\cite{wang2025v1} learns from captions of future
segments and text-only LLM rationales, and further applies
reinforcement learning with multiple-choice answers.
Another line grounds reasoning in future visual states:
TwiFF~\cite{liu2026twiff} reasons with explicit future frames,
FUTURE-L1~\cite{jiang2026futurel1} aligns latent reasoning with
future-frame embeddings before answer-based reinforcement learning with
an LLM judge,
and IVT~\cite{zhu2026ivt} adds next-embedding prediction to
fine-tuning on annotated event descriptions.

Despite this progress, these methods share two limitations.
First, their supervision relies on ground-truth answers, annotated
descriptions, or external large models such as LLM rationale
generators and
judges~\cite{wang2025v1,su2026videocoe,jiang2026futurel1,zhu2026ivt},
which limits scalability.
Second, they treat the future clip as a uniformly valuable target
and fit its full textual or visual
content~\cite{wang2025v1,liu2026twiff,jiang2026futurel1,zhu2026ivt},
yet future information is not equally useful for a given question.
As shown in \cref{fig:teaser}, to predict what to do next, only
\emph{rinse the knife} is event-relevant content information, while
most of the scene (\eg, tap, dish rack, sink) is irrelevant context
information.

To address the first limitation, we adopt an on-policy
self-distillation framework to turn naturally occurring
past--future video pairs into a self-supervised learning signal.
We split each video into a past clip and its future clip (the
\emph{paired future}).
The \emph{student} sees only the past and generates a
reasoning-and-answer response, while a frozen copy of the initial
student that additionally sees the paired future acts as the
\emph{teacher} and scores each token of that response.
As supervision comes from the video's own future, this framework needs
no annotations or external large models and thus scales well.

To address the second limitation, we define a \emph{foresight score}
to teach the model to identify the most valuable future information.
It measures how well a student response matches the true future.
Directly optimizing this score is intractable, as it averages over all
possible futures.
We therefore derive a tractable surrogate objective that provably
lower-bounds the foresight score.
The surrogate interpolates between a \emph{contrastive privilege
advantage}, which compares the teacher's scores of the same response
under the paired true future and an unrelated reference future, and a
\emph{teacher--student distillation advantage} toward the
paired-future teacher.
The joint interpolated objective suppresses shared context information
and amplifies content information specific to the paired future
(\cref{fig:method}).
We refer to the framework combining these two designs as \method.

We evaluate \method on Qwen3-VL and Qwen3.5 across two video foresight
tasks: open-ended prediction and closed-ended planning.
\method improves the foresight reasoning of the base models by up to
14.2 and 15.9 points, respectively, and consistently outperforms Naive OPSD, which does not optimize the foresight score
(\cref{tab:futurel1,tab:egoplan}).
The gains transfer across tasks and largely preserve general video
understanding.
Stronger teachers and more data bring further gains.
Token-level analyses and discussions show that the interpolated objective
suppresses context information and amplifies content information.

Our contributions are:
\begin{itemize}
  \item We propose \method, which is, to our knowledge, the first on-policy self-distillation framework for video foresight reasoning. It turns naturally occurring past--future video pairs into a self-supervised signal, requiring no extra annotations.
  \item We introduce a \emph{foresight score} and derive its tractable lower bound as a token-level training objective, which enables the model to automatically identify and learn the most valuable future information.
  \item Experiments on Qwen3-VL and Qwen3.5 show gains of up to 14.2 and 15.9 points on open-ended prediction and closed-ended planning, with strong cross-task transfer, no compromise in general video understanding, and gains that grow with stronger teachers and more data.
\end{itemize}
\endgroup

\section{Related Work}
\label{sec:related}

\noindent\textbf{Video Foresight Reasoning.}
Early studies on video foresight focus on action anticipation with recurrent and transformer models~\cite{vondrick2016anticipating,furnari2020rolling,girdhar2021anticipative,gong2022futuretransformer,ma2024intention}, future event prediction in natural language~\cite{lei2020futureevent}, and LLM-based long-term action planning~\cite{zhao2024antgpt,kim2024palm}.
With the rise of VLMs, benchmarks such as FutureBench, TwiFF-Bench, EgoPlan-Bench, and SEED-Bench-R1 evaluate future-event reasoning and next-action planning~\cite{wang2025v1,liu2026twiff,chen2023egoplan,chen2025seedr1}, while CausalVQA, FutureOmni, and RIVER extend evaluation to physical, audio-visual, and streaming settings~\cite{foss2025causalvqa,chen2026futureomni,shi2026river}.
Existing training methods fall into three lines:
(i) supervised learning of textual future targets and rationales~\cite{yu2024merlin,wang2025v1};
(ii) modeling explicit or latent future visual states, or predictive video representations~\cite{liu2026twiff,jiang2026futurel1,zhu2026ivt,teeti2023temporaldino,assran2025vjepa2,sun2026mvp};
and (iii) reinforcement learning with answer-, event-, or generation-based rewards~\cite{wang2025v1,su2026videocoe,cheng2025vans}.
These methods rely on annotations or external large models for supervision, or fit the whole future clip uniformly.
In contrast, \method does not fit the future: it uses future clips as privileged evidence to score the student's own responses, without annotations or an extra large LLM as a judge to provide supervision signals.

\noindent\textbf{On-Policy Self-Distillation.}
Learning using privileged information trains a student with information available only during training~\cite{vapnik2009lupi,lopezpaz2016distillation}.
On-policy distillation has a teacher score student-generated trajectories to reduce exposure bias~\cite{agarwal2024gkd}, and OPSD uses the same model, conditioned on privileged context, as its own teacher~\cite{zhao2026opsd}.
Multimodal variants use temporal, high-resolution, or localized visual evidence for video understanding~\cite{li2026videoopd,zhu2026rpopsd,wang2026videoopsd,wang2026clueopsd,venkatraman2026temporal,wu2026streamopd}, and related ideas extend to world-action models and video generators~\cite{fang2026pfd,liu2026opsdv}.
Recent analyses show that privileged context can shift the teacher in task-irrelevant ways, termed privilege illusions~\cite{yu2026dopd,shrestha2026rethinking}; RLCSD contrasts textual hints within outcome-reward RL~\cite{pan2026rlcsd}.
In contrast, \method is, to our knowledge, the first OPSD framework for video foresight reasoning, and differs from prior contrastive distillation in two ways.
(1) \emph{Data}: it contrasts the true future clip of the same video with a future clip from an unrelated video, rather than textual hints~\cite{pan2026rlcsd}.
(2) \emph{Objective}: grounded in a foresight score and its KL-regularized lower bound (\cref{thm:main-bound}), it interpolates contrastive and distillation advantages rather than relying on contrast alone.

\section{Future-OPSD}
\label{sec:method}

\begin{figure*}[t]
  \centering
  \includegraphics[width=\textwidth]{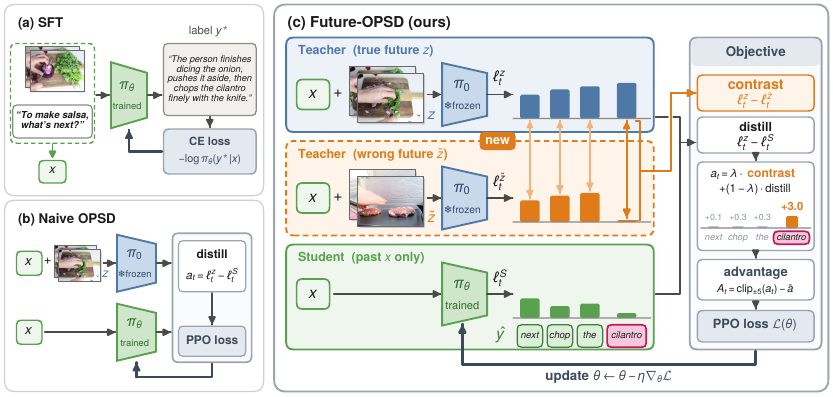}
  \caption{\textbf{Overview of Future-OPSD and comparison with other training paradigms.}
  (a) \textbf{SFT} learns annotated future descriptions $y^\star$ from past observations $x$ using cross-entropy.
  (b) \textbf{Naive OPSD} uses a frozen future-aware teacher $\pi_0$ to score the student's own response $\hat y$, converting the teacher--student log-probability gap into token advantages for PPO.
  (c) \textbf{Future-OPSD (ours)} has a frozen teacher score the same response under the true future $z$ (blue) and a reference future $\tilde z$ from another video (orange).
  The \emph{contrastive privilege advantage} $\ell_t^z-\ell_t^{\tilde z}$ compares the two futures, and the \emph{teacher--student distillation advantage} $\ell_t^z-\ell_t^S$ compares the true-future teacher with the past-only student.
  Their interpolation $a_t$ down-weights context tokens supported by both futures (\emph{next / chop / the}) while keeping content tokens specific to the true future (\emph{cilantro}), helping the student focus on the most valuable future information.
  The advantages are clipped and centered into $A_t$ for PPO, and inference uses only the student.}
  \label{fig:method}
\end{figure*}

We propose \method, whose overview is shown in \cref{fig:method}.
We first set up the problem (\cref{sec:setup}) and define the
foresight score (\cref{sec:foresight}). We then derive a tractable
contrastive surrogate (\cref{sec:contrast}), control its gap with KL
regularization (\cref{sec:bound}), and turn the objective into
token-level PPO updates (\cref{sec:advantage}).

\subsection{Problem Setup and Notation}
\label{sec:setup}

For paired data $(X,Z)\sim p$, $X$ denotes past frames and a question,
and $Z$ the paired future, \ie, the subsequent clip recorded in the same video. The student $\pi_\theta$ generates a
reasoning-and-answer response $Y$ from $X$ alone; the frozen teacher
$\pi_0$ scores it given $(X,Z)$. Thus,
\mbox{$Y\sim\pi_\theta(\cdot\mid X)$} and
\mbox{$p_\theta(y\mid x,z)=\pi_\theta(y\mid x)$}.
This gives $Y\ind Z\mid X$ and the Markov chain $Z\to X\to Y$.
We sample a reference future $\zt\sim p_Z$ from another video,
independently of $(X,Z,Y)$, where $p_Z$ is the future marginal.
Training requires no answer or rationale labels.
Inference uses only the student. We assume finite past, future, and response spaces
and positive teacher probabilities for all allowed responses.
Full assumptions and proofs appear in \cref{app:proof}.

\subsection{Defining the Foresight Score}
\label{sec:foresight}

\noindent\textbf{How to measure video foresight?}
A general prediction may be compatible with many futures.
Inspired by InfoNCE~\cite{oord2018cpc,poole2019variational}, we measure
how specifically the paired future supports a response by comparing
its teacher probability with the average under reference futures.
We define the \textbf{foresight score}
\begin{equation}
  \Phi(\theta)=
  \E_{\substack{(x,z)\sim p\\y\sim\pi_\theta(\cdot\mid x)}}
  \left[
    \log\frac{\pi_0(y\mid x,z)}
    {\E_{\zt\sim p_Z}\pi_0(y\mid x,\zt)}
  \right].
  \label{eq:phi}
\end{equation}
Normalizing the teacher likelihoods defines a verifier distribution
$q(z\mid x,y)\propto p_Z(z)\pi_0(y\mid x,z)$.
Since the teacher is frozen, $\theta$ affects the foresight score only
through the student's sampling distribution $\pi_\theta(y\mid x)$:
different parameters generate different responses, which receive
different relative support.
The score corresponds to the limit of normalized InfoNCE as the
number of reference futures tends to infinity; the connection is
detailed in \cref{app:identity}.

\begin{samepage}
\begin{theorem}[Foresight identity]
\label{thm:main-identity}
Under the conditions in \cref{app:setup}, a past-only student satisfies
\begin{equation}
  \begin{aligned}
  \Phi(\theta)&=I(X;Z)-D(\theta)\leq I(X;Z),\\
  D(\theta)&=\E_{x,y}\KL\!\left(
    p(\cdot\mid x)\,\middle\|\,q(\cdot\mid x,y)\right).
  \end{aligned}
  \label{eq:foresight-identity}
\end{equation}
\end{theorem}
\end{samepage}

Here, $D(\theta)$ is the average KL divergence from the true
conditional future distribution $p(z\mid x)$ to the teacher-defined
verifier $q(z\mid x,y)$.
Since $I(X;Z)$ is fixed by the data, maximizing $\Phi$ is equivalent
to reducing this discrepancy. Training thus favors responses that
better align the verifier with the true future distribution.
The proof is provided in \cref{app:identity}.

However, evaluating $\Phi$ requires the normalizer of $q$, an
expectation over the full future distribution that is generally
intractable to compute exactly. We therefore introduce a surrogate
that can be estimated from sampled reference futures and decomposes
over tokens.

\subsection{Comparing Scores across Futures}
\label{sec:contrast}
\label{sec:psd}
\label{sec:relative-evidence}

\noindent\textbf{How to optimize $\Phi(\theta)$ token by token?}
We compare the teacher's log-probabilities of the same response
under paired and reference futures. Their difference decomposes
over tokens, giving the surrogate:
\begin{equation}
  \Lc(\theta)=\E_{x,z,y,\zt}
  \left[
    \log\pi_0(y\mid x,z)-\log\pi_0(y\mid x,\zt)
  \right].
  \label{eq:linear}
\end{equation}
Both teacher evaluations use the same past, response, and input
format, so shared additive score components cancel.
Log-probabilities add over tokens, enabling token-level supervision.
However, the contrast includes an extra nonnegative Jensen gap:
\begin{equation}
  \begin{aligned}
  \Lc(\theta)&=\Phi(\theta)+\Gamma(\theta),\\
  \Gamma(\theta)&=\E_{x,y}\big[
    \log\E_{\zt}\pi_0(y\mid x,\zt)\\
    &\hspace{4.6em}-\E_{\zt}\log\pi_0(y\mid x,\zt)\big]\geq0.
  \end{aligned}
  \label{eq:gap}
\end{equation}
The student may increase $\Gamma$ by favoring responses with larger
score differences across reference futures, raising the contrastive
score without improving foresight. We therefore regularize this
objective.

\subsection{Controlling the Contrastive Gap}
\label{sec:bound}
\label{sec:cpsd}

\noindent\textbf{How to control the gap between $\Lc$ and $\Phi$?}
We encourage future discrimination while using a KL penalty to limit
the student's deviation from the paired-future teacher.
The sequence-level penalty is
\begin{equation}
  \Rc(\theta)=\E_{x,z}\KL\!\left(
    \pi_\theta(\cdot\mid x)\,\middle\|\,\pi_0(\cdot\mid x,z)\right).
  \label{eq:sequence-kl}
\end{equation}
We maximize the regularized objective
\begin{equation}
  J_\lambda(\theta)=
    \lambda\Lc(\theta)-(1-\lambda)\Rc(\theta),
  \qquad \lambda\in[0,1].
  \label{eq:objective}
\end{equation}
For $0<\lambda<1$, set $\beta=\lambda/(1-\lambda)>0$.
A change-of-measure inequality gives
$\Gamma(\theta)\leq\Rc(\theta)/\beta+C_\beta$
(see \cref{app:certificate} for the derivation and proof).
Substituting this into $\Phi=\Lc-\Gamma$
yields a lower bound.

\begin{samepage}
\begin{theorem}[KL lower bound on the foresight score]
\label{thm:main-bound}
Under the conditions in \cref{app:setup}, fix $0<\lambda<1$ and
$\beta=\lambda/(1-\lambda)$. There is a finite constant $C_\beta$,
depending only on the data, frozen teacher, and $\beta$, such that
\begin{equation}
  \Phi(\theta)\geq\Lc(\theta)-\frac{\Rc(\theta)}{\beta}-C_\beta
  =\frac{J_\lambda(\theta)}{\lambda}-C_\beta.
  \label{eq:bound}
\end{equation}
\end{theorem}
\end{samepage}

For fixed $\lambda$, maximizing $J_\lambda$ maximizes this lower bound.
Since $C_\beta$ is independent of $\theta$, training need not compute
it. A higher bound need not imply a higher $\Phi$.

The weight $\lambda$ balances the contrastive privilege and teacher--student distillation advantages.
At $\lambda=0$, the objective reduces to naive distillation toward
the paired-future teacher; at $\lambda=1$, it retains only the
contrast without teacher regularization, which may lead to training
collapse (\cref{fig:lambda}).
We next derive token-level updates for this sequence objective.

\providecommand{\msd}[2]{#1}
\providecommand{\msdimage}[2]{#1}
\providecommand{\basedelta}[1]{\textcolor{blue}{#1}}
\providecommand{\swatch}[1]{{\setlength{\fboxsep}{0pt}\setlength{\fboxrule}{0.4pt}\fcolorbox{black!50}{#1}{\rule{0pt}{0.7em}\rule{0.7em}{0pt}}}}
\newsavebox{\generalresultstable}
\newsavebox{\generalresultsheader}
\newcommand{\resultsseparator}[1]{%
  \rlap{\hspace{\dimexpr#1\relax}%
    \textcolor{black!30}{\rule[-\dp\generalresultstable]{0.3pt}{%
      \dimexpr\ht\generalresultstable+\dp\generalresultstable
        -\ht\generalresultsheader-\dp\generalresultsheader+\belowrulesep\relax}}}}
\newcommand{\printgeneralresultstable}{%
  \sbox{\generalresultsheader}{%
    \begin{tabular}{@{}c@{}}
      \toprule
      \mbox{}\\
      \cmidrule{1-1}
      \mbox{}\\
      \midrule
    \end{tabular}}%
  \leavevmode
  \resultsseparator{\wd\generalresultstable-28em-15\tabcolsep-0.6pt}%
  \resultsseparator{\wd\generalresultstable-17.5em-9\tabcolsep-0.3pt}%
  \usebox{\generalresultstable}}
\begin{table*}[tp]
\centering
\captionsetup{position=top,skip=0pt}
\caption{\textbf{Open-ended prediction (trained on Future-L1).}
\swatch{seedbg}~Blue cells denote in-domain evaluation.
Abbreviations: Temp.~=~TempCompass, \mbox{T-B.}~=~TemporalBench, \mbox{V-M.}~=~Video-MME, and MVB~=~MVBench.
Bold marks the highest score in each column within each model.
Future-OPSD improves in-domain performance by up to 14.2 points over the base models while preserving general video understanding.}
\label{tab:futurel1}
\label{tab:futurel1-q35}
\footnotesize
\renewcommand{\arraystretch}{0.97}
\settowidth{\dimen0}{Naive OPSD~\cite{zhao2026opsd}}
\settowidth{\dimen2}{Labels}
\settowidth{\dimen4}{\textbf{Qwen3-VL-8B}}
\setlength{\tabcolsep}{\dimexpr(\textwidth-\dimen0-\dimen2-\dimen4-38.5em-0.6pt)/26\relax}
\ifdim\tabcolsep<1pt\setlength{\tabcolsep}{1pt}\fi
\begin{lrbox}{\generalresultstable}
\begin{tabular}{@{} l l c *{3}{>{\columncolor{seedbg}}w{c}{3.5em}} !{\hspace{0.3pt}}
  *{3}{w{c}{3.5em}} !{\hspace{0.3pt}}
  *{5}{w{c}{3.5em}} @{}}
\toprule
& & & \multicolumn{3}{>{\columncolor{seedbg}}c}{TwiFF-Bench} & \multicolumn{3}{!{\hspace{0.3pt}}c}{SEED-Bench-R1} & \multicolumn{5}{!{\hspace{0.3pt}}c}{\makebox[0pt]{General Video Understanding}} \\
\cmidrule(lr){4-6}\cmidrule(lr){7-9}\cmidrule(lr){10-14}
Model & Method & Labels & CoT & Ans. & Avg & L1 & L2 & L3 & Temp. & T-B. & V-M. & MVB & Avg \\
\midrule
\multirow{5}{*}{\textbf{Qwen3-VL-8B}} & Base & -- & 58.6 & 48.7 & 53.6 & 45.4 & 47.5 & 39.7 & 78.3 & 18.6 & 61.7 & 65.5 & 56.0 \\
& SFT~\cite{wang2025v1} & \cmark & 61.3 & 55.7 & 58.5 & 33.8 & 35.1 & 33.4 & 71.8 & 16.3 & 54.1 & 57.0 & 49.8 \\
& Naive OPSD~\cite{zhao2026opsd} & -- & 66.0 & 56.4 & 61.2 & 45.9 & 47.3 & 38.7 & 79.0 & \textbf{20.9} & 62.0 & 66.2 & 57.0 \\
& \textbf{Future-OPSD} & -- & \textbf{71.9} & \textbf{63.5} & \textbf{67.8} & \textbf{46.1} & \textbf{48.3} & \textbf{40.0} & \textbf{79.6} & 20.4 & \textbf{63.0} & \textbf{67.0} & \textbf{57.5} \\
\cmidrule[0.3pt](l){2-14}
& \basedelta{$\Delta$ vs. Base} & \basedelta{--} & \basedelta{$+13.3$} & \basedelta{$+14.8$} & \basedelta{$+14.2$} & \basedelta{$+0.7$} & \basedelta{$+0.8$} & \basedelta{$+0.3$} & \basedelta{$+1.3$} & \basedelta{$+1.8$} & \basedelta{$+1.3$} & \basedelta{$+1.5$} & \basedelta{$+1.5$} \\
\midrule[0.8pt]
\multirow{5}{*}{\textbf{Qwen3.5-9B}} & Base & -- & 62.4 & 54.5 & 58.5 & 44.3 & 44.9 & 37.9 & \textbf{79.9} & 21.4 & 61.2 & 65.6 & 57.0 \\
& SFT~\cite{wang2025v1} & \cmark & 63.5 & 57.4 & 60.5 & 36.4 & 37.6 & 31.3 & 73.7 & 19.4 & 56.4 & 48.2 & 49.4 \\
& Naive OPSD~\cite{zhao2026opsd} & -- & 66.2 & 59.1 & 62.7 & 43.3 & 42.4 & 35.2 & \textbf{79.9} & 21.4 & 60.7 & 65.0 & 56.7 \\
& \textbf{Future-OPSD} & -- & \textbf{70.4} & \textbf{66.1} & \textbf{68.2} & \textbf{45.4} & \textbf{47.1} & \textbf{38.0} & 79.3 & \textbf{21.5} & \textbf{63.1} & \textbf{65.9} & \textbf{57.2} \\
\cmidrule[0.3pt](l){2-14}
& \basedelta{$\Delta$ vs. Base} & \basedelta{--} & \basedelta{$+8.0$} & \basedelta{$+11.6$} & \basedelta{$+9.7$} & \basedelta{$+1.1$} & \basedelta{$+2.2$} & \basedelta{$+0.1$} & \basedelta{$-0.6$} & \basedelta{$+0.1$} & \basedelta{$+1.9$} & \basedelta{$+0.3$} & \basedelta{$+0.2$} \\
\bottomrule
\end{tabular}
\end{lrbox}
\printgeneralresultstable
\end{table*}

\begin{table*}[tp]
\centering
\captionsetup{position=top,skip=0pt}
\caption{\textbf{Closed-ended planning (trained on EgoPlan-IT).}
\swatch{twiffbg}~Yellow cells denote in-domain evaluation.
Abbreviations: Temp.~=~TempCompass, \mbox{T-B.}~=~TemporalBench, \mbox{V-M.}~=~Video-MME, and MVB~=~MVBench.
Bold marks the highest score in each column within each model.
Future-OPSD improves in-domain performance by up to 15.9 points over the base models while preserving general video understanding.}
\label{tab:egoplan-q3vl}
\label{tab:egoplan}
\label{tab:egoplan-q35}
\footnotesize
\renewcommand{\arraystretch}{0.96}
\settowidth{\dimen0}{Naive OPSD~\cite{zhao2026opsd}}
\settowidth{\dimen2}{Labels}
\settowidth{\dimen4}{\textbf{Qwen3-VL-8B}}
\setlength{\tabcolsep}{\dimexpr(\textwidth-\dimen0-\dimen2-\dimen4-38.5em-0.6pt)/26\relax}
\ifdim\tabcolsep<1pt\setlength{\tabcolsep}{1pt}\fi
\begin{lrbox}{\generalresultstable}
\begin{tabular}{@{} l l c *{3}{>{\columncolor{twiffbg}}w{c}{3.5em}} !{\hspace{0.3pt}}
  *{3}{w{c}{3.5em}} !{\hspace{0.3pt}}
  *{5}{w{c}{3.5em}} @{}}
\toprule
& & & \multicolumn{3}{>{\columncolor{twiffbg}}c}{SEED-Bench-R1} & \multicolumn{3}{!{\hspace{0.3pt}}c}{TwiFF-Bench} & \multicolumn{5}{!{\hspace{0.3pt}}c}{\makebox[0pt]{General Video Understanding}} \\
\cmidrule(lr){4-6}\cmidrule(lr){7-9}\cmidrule(lr){10-14}
Model & Method & Labels & L1 & L2 & L3 & CoT & Ans. & Avg & Temp. & T-B. & V-M. & MVB & Avg \\
\midrule
\multirow{6}{*}{\textbf{Qwen3-VL-8B}} & Base & -- & 45.4 & 47.5 & 39.7 & 58.6 & 48.7 & 53.6 & 78.3 & 18.6 & 61.7 & 65.5 & 56.0 \\
& SFT~\cite{wang2025v1} & \cmark & 47.4 & 48.6 & 42.9 & 50.0 & 45.7 & 47.8 & 66.9 & 16.3 & 55.4 & 55.4 & 48.5 \\
& TTRL~\cite{ttrl} & -- & 45.8 & 50.3 & 42.9 & 60.7 & 50.1 & 55.4 & \textbf{78.9} & 19.8 & 61.5 & \textbf{66.7} & \textbf{56.7} \\
& Naive OPSD~\cite{zhao2026opsd} & -- & 50.1 & 51.7 & 44.9 & 60.9 & 47.7 & 54.3 & 78.6 & \textbf{20.1} & 61.3 & 65.7 & 56.4 \\
& \textbf{Future-OPSD} & -- & \msdimage{\textbf{54.3}}{0.4} & \msdimage{\textbf{55.8}}{0.7} & \msdimage{\textbf{50.6}}{0.9} & \msdimage{\textbf{63.8}}{1.3} & \msdimage{\textbf{53.1}}{1.8} & \msdimage{\textbf{58.5}}{1.6} & \msdimage{78.8}{0.3} & \msdimage{19.3}{0.6} & \msdimage{\textbf{62.7}}{0.3} & \msdimage{65.1}{0.4} & \msdimage{56.5}{0.3} \\
\cmidrule[0.3pt](l){2-14}
& \basedelta{$\Delta$ vs. Base} & \basedelta{--} & \basedelta{$+8.9$} & \basedelta{$+8.3$} & \basedelta{$+10.9$} & \basedelta{$+5.2$} & \basedelta{$+4.4$} & \basedelta{$+4.9$} & \basedelta{$+0.5$} & \basedelta{$+0.7$} & \basedelta{$+1.0$} & \basedelta{$-0.4$} & \basedelta{$+0.5$} \\
\midrule[0.8pt]
\multirow{6}{*}{\textbf{Qwen3.5-9B}} & Base & -- & 44.3 & 44.9 & 37.9 & 62.4 & 54.5 & 58.5 & 79.9 & 21.4 & \textbf{61.2} & \textbf{65.6} & 57.0 \\
& SFT~\cite{wang2025v1} & \cmark & 47.5 & 50.1 & 44.9 & 53.1 & 48.1 & 50.6 & 73.5 & 17.4 & 57.0 & 57.2 & 51.3 \\
& TTRL~\cite{ttrl} & -- & 27.3 & 29.7 & 26.4 & 53.9 & 54.6 & 54.2 & 75.7 & 20.3 & 56.0 & 61.0 & 53.3 \\
& Naive OPSD~\cite{zhao2026opsd} & -- & 48.2 & 49.9 & 40.7 & 58.5 & 53.1 & 55.8 & 79.2 & 21.7 & 59.4 & 64.6 & 56.2 \\
& \textbf{Future-OPSD} & -- & \msdimage{\textbf{58.4}}{0.4} & \msdimage{\textbf{59.7}}{1.5} & \msdimage{\textbf{53.8}}{0.5} & \msdimage{\textbf{63.7}}{1.8} & \msdimage{\textbf{57.9}}{1.5} & \msdimage{\textbf{60.8}}{1.6} & \msdimage{\textbf{81.4}}{0.2} & \msdimage{\textbf{21.9}}{0.4} & \msdimage{\textbf{61.2}}{0.7} & \msdimage{65.0}{0.1} & \msdimage{\textbf{57.4}}{0.1} \\
\cmidrule[0.3pt](l){2-14}
& \basedelta{$\Delta$ vs. Base} & \basedelta{--} & \basedelta{$+14.1$} & \basedelta{$+14.8$} & \basedelta{$+15.9$} & \basedelta{$+1.3$} & \basedelta{$+3.4$} & \basedelta{$+2.3$} & \basedelta{$+1.5$} & \basedelta{$+0.5$} & \basedelta{$0.0$} & \basedelta{$-0.6$} & \basedelta{$+0.4$} \\
\bottomrule
\end{tabular}
\end{lrbox}
\printgeneralresultstable
\end{table*}

\subsection{Updating the Student Policy}
\label{sec:advantage}
\label{sec:optimization}

\noindent\textbf{How to turn $J_\lambda$ into a practical update?}
We use immediate token advantages and clipped PPO updates.
At each iteration, we fix a rollout policy $\pi_{\mathrm{old}}$,
sample responses, and draw one reference future per response.
Let $h_t=(x,y_{<t})$,
$\ell_t^u=\log\pi_0(y_t\mid h_t,u)$ for $u\in\{z,\zt\}$, and
$\ell_t^S=\log\pi_{\mathrm{old}}(y_t\mid h_t)$.
We approximate the policy gradient by ignoring later rewards and using
each token's immediate reward under $\pi_{\mathrm{old}}$ as its advantage
(\cref{app:advantage}):
\begin{equation}
  a_t=\lambda\underbrace{(\ell_t^z-\ell_t^{\zt})}_{\text{contrastive privilege}}
  +(1-\lambda)\underbrace{(\ell_t^z-\ell_t^S)}_{\text{teacher--student distillation}}.
  \label{eq:cpsd}
\end{equation}
One independent reference gives an unbiased estimate of the
reference-averaged contrast before clipping. The two terms provide
the contrastive privilege advantage and the teacher--student
distillation advantage, respectively.

We clip advantages to $[-c,c]$ and center them over the set
$\mathcal T$ of valid batch-token pairs $(i,t)$, where $i$ indexes
responses:
\begin{equation}
  \begin{aligned}
  \bar a&=\frac{1}{|\mathcal T|}
    \sum_{(i,t)\in\mathcal T}\clip(a_{i,t},-c,c),\\
  A_{i,t}&=\clip(a_{i,t},-c,c)-\bar a.
  \end{aligned}
  \label{eq:baseline}
\end{equation}
Let $\rho_{i,t}=\pi_\theta(y_{i,t}\mid h_{i,t})/
\pi_{\mathrm{old}}(y_{i,t}\mid h_{i,t})$, with
$h_{i,t}=(x_i,y_{i,<t})$, and
$\bar\rho_{i,t}=\clip(\rho_{i,t},1-\epsilon_l,1+\epsilon_h)$.
Holding $A_{i,t}$ fixed, we minimize the token-averaged PPO
loss~\cite{schulman2017ppo}:
\begin{equation}
  \mathcal L_{\mathrm{PPO}}(\theta)
  =-\frac{1}{|\mathcal T|}\sum_{(i,t)\in\mathcal T}
  \min\!\big(\rho_{i,t}A_{i,t},\bar\rho_{i,t}A_{i,t}\big).
  \label{eq:loss}
\end{equation}
Here, $\epsilon_l$ and $\epsilon_h$ set the clipping range.
Only the student is updated; the teacher, rollout policy, and
advantages remain fixed. Since $a_t$ includes the KL penalty,
no separate KL loss is needed.
The algorithm flowchart and
pseudocode are provided in \cref{app:algorithm}.

\section{Experiments}
\label{sec:experiments}
\raggedbottom

We organize our experiments around three research questions (RQs) and three discussions (DSs).
\textbf{RQ1:} Does Future-OPSD improve VLMs' video foresight reasoning?
\textbf{RQ2:} Do the performance gains persist with stronger teachers or more training data?
\textbf{RQ3:} How does Future-OPSD work?
\textbf{DS1:} How does Future-OPSD compare with GRPO-based methods?
\textbf{DS2:} Which design choices in Future-OPSD matter?
\textbf{DS3:} What is the computational cost?

\subsection{Main Results (RQ1)}
\label{sec:main-results}

\paragraph{Models and datasets.}
We use two open VLM families, Qwen3-VL~\cite{bai2025qwen3vl} and Qwen3.5~\cite{qwen2026qwen35}, with Qwen3-VL-8B-Instruct and Qwen3.5-9B as the students in RQ1.
Both use reason-then-answer prompting, with native thinking disabled for Qwen3.5-9B.
We consider two video foresight tasks:
(a)~\emph{Open-ended prediction}: we train on Future-L1~\cite{jiang2026futurel1} (50K clips, one open-ended question per clip) and evaluate in-domain on TwiFF-Bench~\cite{liu2026twiff} (1,078 questions).
(b)~\emph{Closed-ended planning}: we train on EgoPlan-IT~\cite{chen2023egoplan} (43,294 four-way next-action questions on 30.5K progress clips from 610 EPIC-KITCHENS-100 videos~\cite{damen2022epic}) and evaluate in-domain on SEED-Bench-R1~\cite{chen2025seedr1} (L1--L3).

\paragraph{Training and evaluation.}
We fine-tune all student parameters for 100 steps (256 queries, one rollout each), with learning rate $10^{-6}$, temperature $1.0$, $c=5$, $\lambda=0.75$, and no separate KL loss (\cref{sec:cpsd}).
Teachers remain frozen at initialization and receive up to three future frames; reference futures use the same format from other videos.
Following Future-L1~\cite{jiang2026futurel1}, a Qwen3.6-27B judge scores open-ended reasoning and answers separately on a 1--5 scale, rescaled to 0--100; planning uses A--D accuracy (\%).
We also report accuracy (\%) on TempCompass~\cite{liu2024tempcompass}, TemporalBench~\cite{cai2024temporalbench}, Video-MME~\cite{fu2024videomme}, and MVBench~\cite{li2024mvbench} under standard settings.

\paragraph{Baselines.}
(i) \emph{Base}: pretrained models without task-specific fine-tuning.
(ii) \emph{SFT}~\cite{wang2025v1}: cross-entropy prediction of future states and answers from current observations, using textual targets generated from future frames by a stronger external model.
(iii) \emph{Naive OPSD}~\cite{zhao2026opsd}: token-level distillation from a frozen teacher with future context to a past-only student, without future contrast.
(iv) \emph{TTRL}~\cite{ttrl}: label-free RL with majority-vote pseudo-rewards from eight rollouts (closed-ended planning only).
\label{sec:exp-setup}

\textbf{Future-OPSD improves video foresight reasoning in both open-ended prediction and closed-ended planning while preserving general video understanding.}
\Cref{tab:futurel1} evaluates Qwen3-VL-8B and Qwen3.5-9B trained on Future-L1, using TwiFF-Bench for open-ended prediction.
\Cref{tab:egoplan-q3vl} uses the same students trained on EgoPlan-IT, with SEED-Bench-R1 for closed-ended planning.
Both show consistent gains over Base and Naive OPSD without answer labels, spanning reasoning and answer quality as well as planning across all three generalization levels.
The gains also transfer across foresight tasks while retaining general video understanding.

\subsection{Teacher and Data Scaling (RQ2)}
\label{sec:datascale-q3vl2b}
\raggedbottom

\textbf{Using stronger VLM teachers further improves video foresight reasoning.}
We train on EgoPlan-IT in two settings: (i) a Qwen3-VL-2B-Instruct student with a frozen Qwen3-VL-4B teacher, and (ii) a Qwen3-VL-4B student with a frozen Qwen3-VL-8B teacher.
As all else is fixed, the gain of Future-OPD over Future-OPSD comes from the stronger teacher.
It improves in-domain planning in both settings and transfers to open-ended prediction (\cref{tab:q3vl-egoplan}).

\begin{table}[!t]
\centering
\renewcommand{\arraystretch}{0.92}
\caption{\textbf{Teacher scaling on EgoPlan-IT.}
OPSD uses a frozen copy of the student as teacher; OPD uses a larger frozen teacher (Qwen3-VL-4B for the 2B student, Qwen3-VL-8B for the 4B student).
Stronger teachers improve planning and open-ended prediction, and future contrast outperforms naive distillation.}
\label{tab:q3vl-egoplan}
\label{tab:opd}
\small
\setlength{\tabcolsep}{1pt}
\begin{tabular*}{\linewidth}{@{\extracolsep{\fill}} l ccccc @{}}
\toprule
& \multicolumn{4}{c}{SEED-Bench-R1 (\%)} & TwiFF \\
\cmidrule(lr){2-5}
Method & L1 & L2 & L3 & Avg & Avg \\
\midrule
\multicolumn{6}{l}{\textit{Qwen3-VL-2B-Instruct student}} \\
Base & 34.7 & 36.9 & 30.0 & 33.9 & 34.0 \\
SFT & 39.8 & 42.1 & 37.9 & 39.9 & 40.3 \\
\midrule
\rowcolor{black!6}
\multicolumn{6}{l}{\textbf{OPSD} (frozen 2B as the teacher)} \\
Naive OPSD & 37.0 & 38.5 & 33.8 & 36.4 & 22.6 \\
\textbf{Future-OPSD} & 40.4 & 44.0 & 37.9 & 40.8 & 40.7 \\
\basedelta{$\Delta$ vs. Base} & \basedelta{$+5.7$} & \basedelta{$+7.1$} & \basedelta{$+7.9$} & \basedelta{$+6.9$} & \basedelta{$+6.7$} \\
\midrule
\rowcolor{black!6}
\multicolumn{6}{l}{\textbf{OPD} (frozen Qwen3-VL-4B as the teacher)} \\
Naive OPD & 44.4 & 48.9 & 36.9 & 43.4 & 34.3 \\
\textbf{Future-OPD} & \textbf{48.4} & \textbf{51.9} & \textbf{44.4} & \textbf{48.2} & \textbf{41.3} \\
\basedelta{$\Delta$ vs. Base} & \basedelta{$+13.7$} & \basedelta{$+15.0$} & \basedelta{$+14.4$} & \basedelta{$+14.3$} & \basedelta{$+7.3$} \\
\midrule
\multicolumn{6}{l}{\textit{Qwen3-VL-4B student}} \\
Base & 40.2 & 42.0 & 34.1 & 38.8 & 50.5 \\
SFT & 41.2 & 41.3 & 41.8 & 41.4 & 47.5 \\
\midrule
\rowcolor{black!6}
\multicolumn{6}{l}{\textbf{OPSD} (frozen 4B as the teacher)} \\
Naive OPSD & 47.2 & 49.4 & 40.3 & 45.6 & 51.1 \\
\textbf{Future-OPSD} & 50.6 & 51.2 & 45.0 & 48.9 & 54.1 \\
\basedelta{$\Delta$ vs. Base} & \basedelta{$+10.4$} & \basedelta{$+9.2$} & \basedelta{$+10.9$} & \basedelta{$+10.1$} & \basedelta{$+3.6$} \\
\midrule
\rowcolor{black!6}
\multicolumn{6}{l}{\textbf{OPD} (frozen Qwen3-VL-8B as the teacher)} \\
Naive OPD & 48.1 & 50.1 & 42.2 & 46.8 & 55.7 \\
\textbf{Future-OPD} & \textbf{51.4} & \textbf{54.2} & \textbf{46.7} & \textbf{50.8} & \textbf{58.0} \\
\basedelta{$\Delta$ vs. Base} & \basedelta{$+11.2$} & \basedelta{$+12.2$} & \basedelta{$+12.6$} & \basedelta{$+12.0$} & \basedelta{$+7.5$} \\
\bottomrule
\end{tabular*}
\end{table}

\begin{figure}[!t]
  \centering
  \includegraphics[width=0.85\columnwidth]{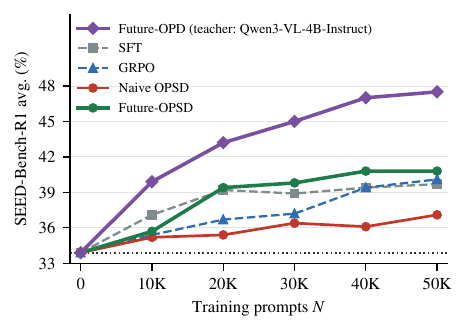}
  \caption{\textbf{Data scaling on EgoPlan-IT (Qwen3-VL-2B).}
  We compare SFT and GRPO trained with answer labels against Naive OPSD, Future-OPSD, and its stronger-teacher variant (frozen Qwen3-VL-4B-Instruct).
  Future-OPSD scales better than naive distillation and surpasses SFT and GRPO at larger data scales, with further gains from a stronger teacher.}
  \label{fig:datascale-q3vl2b}
\end{figure}

\textbf{Future-OPSD improves video foresight reasoning with more training data, and a stronger teacher further amplifies the gains.}
We train Qwen3-VL-2B-Instruct on increasing amounts of EgoPlan-IT data, up to 50K prompts, with either a frozen self-teacher or a frozen Qwen3-VL-4B-Instruct teacher.
\Cref{fig:datascale-q3vl2b} shows that Future-OPSD scales better than Naive OPSD and overtakes SFT and GRPO at larger scales without answer labels, while the stronger teacher enables more sustained gains.

\begin{figure*}[!t]
\centering
\includegraphics[width=\textwidth]{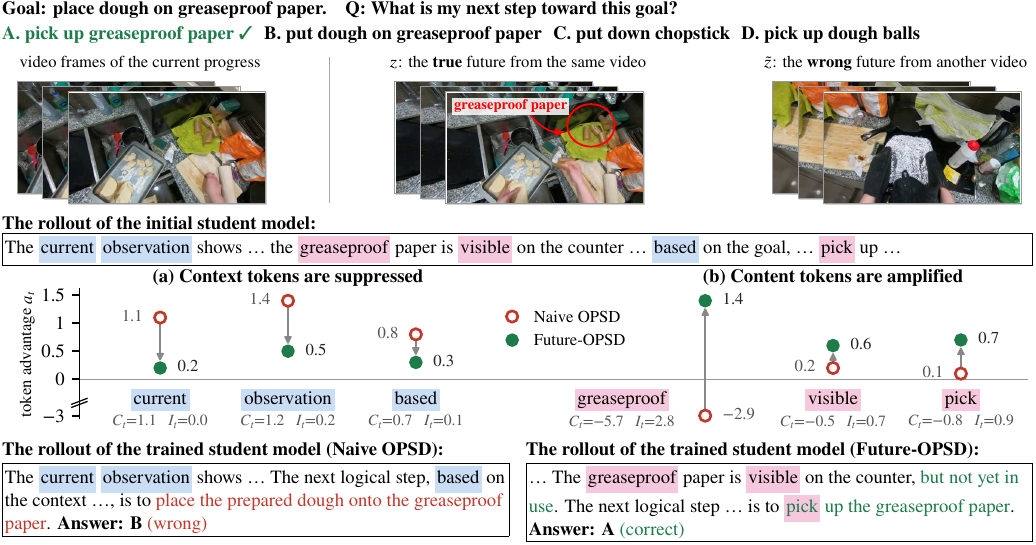}
\caption{\textbf{Case study: why Future-OPSD improves video foresight reasoning.}
On an EgoPlan-IT example with Qwen3-VL-8B, we compare the initial token advantages and trained predictions of Naive OPSD and Future-OPSD.
A frozen teacher scores the same initial rollout under the paired future $z$ and a reference future $\tilde z$ from another video, giving token log-probabilities $\ell_t^z$ and $\ell_t^{\tilde z}$; $\ell_t^S$ is the past-only student's log-probability.
The context and content components are $C_t=\ell_t^{\tilde z}-\ell_t^S$ and $I_t=\ell_t^z-\ell_t^{\tilde z}$.
Before clipping and centering, Naive OPSD uses $a_t=I_t+C_t$, while Future-OPSD uses $a_t=I_t+(1-\lambda)C_t$ with $\lambda=0.75$.
Plotted values are initial advantages, with $C_t$ and $I_t$ below each token.
Future-OPSD suppresses context information and amplifies decision-relevant content information, yielding the correct next-action prediction.}
\label{fig:context}
\end{figure*}

\subsection{How Does Future-OPSD Work? (RQ3)}
\label{sec:why}
\label{sec:context-effect}

\textbf{Future-OPSD amplifies supervision on content information and suppresses supervision on context information.}
In \cref{fig:context}, Naive OPSD reinforces generic descriptions such as ``observation'' while penalizing the decision-relevant token ``greaseproof.''
Future-OPSD redirects supervision toward evidence for the immediate next action.
After training, it correctly predicts picking up the paper (A), whereas Naive OPSD jumps to placing dough on it (B).
\enlargethispage{\baselineskip}
This case illustrates how future contrast helps distinguish the next step from the overall goal.

\begin{table}[!t]
\centering
\caption{\textbf{GRPO versus Future-OPSD.}
Models are trained on Future-L1 and evaluated on TwiFF-Bench and SEED-Bench-R1 (mean of L1--L3).
GRPO uses the student's own base model as an LLM judge to provide rewards.
\swatch{seedbg}~Blue cells indicate in-domain evaluation.}
\label{tab:grpo-comparison}
\label{tab:grpo-futurel1}
\label{tab:grpo-egoplan}
\footnotesize
\setlength{\tabcolsep}{1pt}
\begin{tabular*}{\linewidth}{@{\extracolsep{\fill}} lll cc @{}}
\toprule
Train data & Model & Method & \shortstack{TwiFF\\Avg} & \shortstack{SEED\\Avg} \\
\midrule
\multirow{8}{*}{Future-L1}
& \multirow{2}{*}{\textbf{Qwen3-VL-2B}} & GRPO & \cellcolor{seedbg}52.1 & 26.5 \\
& & \textbf{Future-OPSD} & \cellcolor{seedbg}\textbf{59.2} & \textbf{35.5} \\
\cmidrule(lr){2-5}
& \multirow{2}{*}{\textbf{Qwen3-VL-8B}} & GRPO & \cellcolor{seedbg}66.1 & 41.3 \\
& & \textbf{Future-OPSD} & \cellcolor{seedbg}\textbf{67.8} & \textbf{44.8} \\
\cmidrule(lr){2-5}
& \multirow{2}{*}{\textbf{Qwen3.5-2B}} & GRPO & \cellcolor{seedbg}52.7 & 32.8 \\
& & \textbf{Future-OPSD} & \cellcolor{seedbg}\textbf{55.9} & \textbf{35.2} \\
\cmidrule(lr){2-5}
& \multirow{2}{*}{\textbf{Qwen3.5-9B}} & GRPO & \cellcolor{seedbg}64.4 & 41.6 \\
& & \textbf{Future-OPSD} & \cellcolor{seedbg}\textbf{68.2} & \textbf{43.5} \\
\bottomrule
\end{tabular*}
\end{table}

\begin{table}[!t]
\centering
\caption{\textbf{Text versus image privilege on Future-L1.}
Teachers receive future text or frames during Future-L1 training; students are evaluated on TwiFF-Bench.
Blue parentheses show gains over the paired Naive OPSD row.
Future frames outperform text, and Future-OPSD surpasses Naive OPSD with either modality.}
\label{tab:privilege-futurel1}
\footnotesize
\newcommand{\privgain}[1]{\,\textcolor{blue}{(+#1)}}
\setlength{\tabcolsep}{2pt}
\begin{tabular*}{\linewidth}{@{\extracolsep{\fill}} l lll @{}}
\toprule
Method & CoT & Ans. & Avg \\
\midrule
\rowcolor{black!6}
\multicolumn{4}{l}{\textbf{Qwen3-VL-8B}} \\
\midrule
Naive OPSD ($z=\text{text}$) & 63.4 & 52.6 & 58.0 \\
\textbf{Future-OPSD} ($z=\text{text}$) & \msd{67.9}{1.2}\privgain{4.5} & \msd{58.5}{1.2}\privgain{5.9} & \msd{63.2}{1.2}\privgain{5.2} \\
Naive OPSD ($z=\text{image}$) & 66.0 & 56.4 & 61.2 \\
\textbf{Future-OPSD} ($z=\text{image}$) & \textbf{71.9}\privgain{5.9} & \textbf{63.5}\privgain{7.1} & \textbf{67.8}\privgain{6.6} \\
\midrule
\rowcolor{black!6}
\multicolumn{4}{l}{\textbf{Qwen3.5-9B}} \\
\midrule
Naive OPSD ($z=\text{text}$) & 64.4 & 59.0 & 61.7 \\
\textbf{Future-OPSD} ($z=\text{text}$) & \msd{69.6}{0.4}\privgain{5.2} & \msd{62.1}{0.5}\privgain{3.1} & \msd{65.8}{0.2}\privgain{4.1} \\
Naive OPSD ($z=\text{image}$) & 66.2 & 59.1 & 62.7 \\
\textbf{Future-OPSD} ($z=\text{image}$) & \textbf{70.4}\privgain{4.2} & \textbf{66.1}\privgain{7.0} & \textbf{68.2}\privgain{5.5} \\
\bottomrule
\end{tabular*}
\end{table}

\subsection{Comparison with GRPO (DS1)}
\label{sec:grpo-comparison}

\textbf{GRPO is less effective than Future-OPSD for open-ended prediction.}
We compare Qwen3-VL (2B/8B) and Qwen3.5 (2B/9B) trained on Future-L1.
GRPO~\cite{shao2024deepseekmath} runs for 100 steps with eight rollouts per query, using the student's frozen base model as its reward judge.
It underperforms on both in-domain TwiFF-Bench and cross-domain SEED-Bench-R1 (\cref{tab:grpo-comparison}).
We see two possible causes: (1) open-ended answers lack verifiable rewards, and the LLM judge cannot provide accurate reward signals; (2) even with accurate rewards, reinforcing only the student's own rollouts may not yield cross-domain generalization.

\subsection{Which Design Matters? (DS2)}
\label{sec:design}
\label{sec:ablations}
\label{sec:ablation}

\textbf{Images outperform text.}
On Future-L1, $z=\text{text}$ comprises descriptions generated by Qwen3.5-27B from the frames used as $z=\text{image}$.
Both methods achieve higher scores with images (\cref{tab:privilege-futurel1}).
A possible explanation is that captions compress the future before teacher scoring,
whereas frames let the teacher select event evidence conditioned on
the prediction question. The Qwen3-VL-8B image advantage is consistent
with retaining cues omitted by question-agnostic captions, although
the score comparison alone does not establish this mechanism.

\begin{figure}[H]
\centering
\vspace{-6pt}
\includegraphics[width=\columnwidth]{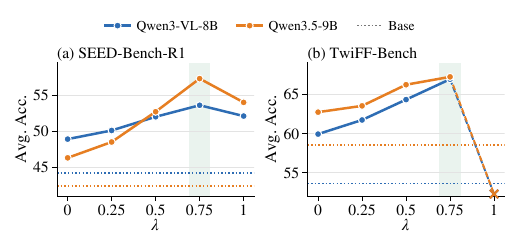}
\caption{\textbf{Search over the mixing weight $\lambda$.}
We sweep $\lambda\in\{0,0.25,0.5,0.75,1\}$ for Qwen3-VL-8B and Qwen3.5-9B on both tasks: (a) EgoPlan-IT training with SEED-Bench-R1 evaluation; (b) Future-L1 training with TwiFF-Bench evaluation.
Among the tested values, $\lambda=0.75$ achieves the best overall performance, balancing contrastive learning and distillation.}
\label{fig:lambda}
\vspace{-6pt}
\end{figure}

\textbf{Our search identifies $\lambda=0.75$ as the best overall setting.}
We sweep $\lambda\in\{0,0.25,0.5,0.75,1\}$ for both students on EgoPlan-IT and Future-L1 (\cref{fig:lambda}).
Combining contrastive and distillation signals improves performance, while pure contrast ($\lambda=1$) collapses training on Future-L1.
This is consistent with \cref{sec:cpsd}: at $\lambda=1$, the KL anchor to the frozen future-conditioned teacher disappears, allowing policy drift and a larger surrogate gap without improved foresight.

\begin{figure}[!t]
\centering
\includegraphics[width=0.75\columnwidth]{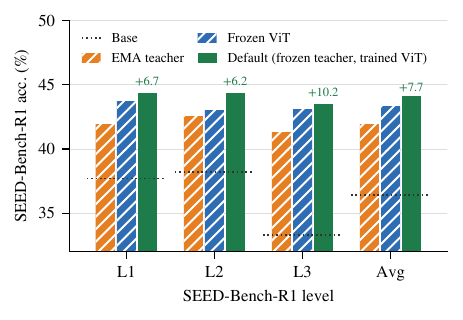}
\caption{\textbf{Teacher weights and vision encoder} (Qwen3.5-2B, EgoPlan-IT).
Each variant changes one setting of the default (frozen teacher, trained ViT).
}
\label{fig:teacher-vit}
\vspace{-12pt}
\end{figure}

\textbf{A frozen teacher and a trainable vision encoder work best.}
We vary which components are updated during training: the teacher (frozen at initialization or updated by EMA) and the student's vision encoder (trained or frozen).
Keeping the teacher frozen without EMA works best, suggesting that a fixed teacher gives a more stable reference, and jointly training the vision encoder further improves SEED-Bench-R1 accuracy (\cref{fig:teacher-vit}).

\subsection{What Is the Computational Cost? (DS3)}
\label{sec:cost}

\textbf{Future-OPSD adds only constant-factor overhead to Naive OPSD.}
Let $U$ denote student forward/backward optimization, $R$ rollout generation and shared policy scoring, and $F$ one frozen-teacher forward pass.
For matched update budgets, SFT costs $U$, Naive OPSD $R+U+F$, and Future-OPSD $R+U+2F$ per query.
The additional scoring pass reuses the teacher without extra generation, backpropagation, or optimizer states, preserving asymptotic complexity.
\Cref{tab:cost} shows approximately 19--33\% more FLOPs and 8--15\% more training time than Naive OPSD with similar memory, consistent with bounded overhead.

\begin{table}[htbp]
\centering
\caption{\textbf{Training cost of Qwen3-VL-8B (8 B200 GPUs).}
FLOPs in EFLOPs; Mem is the peak allocated memory averaged over the eight GPUs (GiB); Time is wall-clock hours (node-normalized for Future-OPSD on Future-L1).}
\label{tab:cost}
\footnotesize
\setlength{\tabcolsep}{2pt}
\begin{tabular*}{\columnwidth}{@{\extracolsep{\fill}} ll cccc @{}}
\toprule
Data & Method & FLOPs & Mem & Time (h) & Score \\
\midrule
\multirow{3}{*}{EgoPlan-IT} & SFT & 4.0 & 63.4 & 0.62 & 44.3 \\
 & Naive OPSD & 9.4 & 53.6 & 3.28 & 48.9 \\
 & \textbf{Future-OPSD} & 11.2 & 54.0 & 3.77 & \textbf{53.6} \\
\midrule
\multirow{3}{*}{Future-L1} & SFT & 1.4 & 68.4 & 0.20 & 57.3 \\
 & Naive OPSD & 4.3 & 43.0 & 1.60 & 61.2 \\
 & \textbf{Future-OPSD} & 5.7 & 42.5 & 1.73 & \textbf{66.9} \\
\bottomrule
\end{tabular*}
\end{table}
\flushbottom

\section{Conclusion}
\label{sec:conclusion}

Existing methods for video foresight reasoning depend on annotations or external large models, and learn from the entire future clip even though not all of it is equally important.
To address both limitations, we proposed \method.
It adopts on-policy self-distillation to turn naturally occurring past--future video pairs into a self-supervised signal, and optimizes a tractable lower bound of a foresight score that interpolates a contrastive privilege advantage with a teacher--student distillation advantage, helping the student focus on the most valuable future information.
Experiments on Qwen3-VL and Qwen3.5 show that \method improves video foresight reasoning without compromising general video understanding, and scales with stronger teachers and more data.
Token-level analyses confirm that it amplifies content information and suppresses context information.
Although our conclusions are limited to the datasets and models studied here, we believe our findings provide useful guidance for improving VLM foresight reasoning.

{
    \small
    \bibliographystyle{ieeenat_fullname}
    \bibliography{main}
}
\end{document}